\pdfoutput=1
\documentclass[runningheads]{llncs}
\usepackage{graphicx}
\usepackage[colorlinks,linkcolor=blue]{hyperref}
\usepackage{multirow}
\usepackage{booktabs} % 用于表格处理，头部线条加粗
\begin{document}
\title{A Dual-Attention Neural Network for Pun Location and Using Pun-Gloss Pairs for Interpretation}
\titlerunning{DANN for Pun Location and Using Pun-Gloss Pairs for Interpretation}

\author{
Shen Liu \inst{1},
Meirong Ma \inst{2},
Hao Yuan \inst{2},
Jianchao Zhu \inst{2},
Yuanbin Wu\inst{1},
Man Lan\inst{1*} 
}

%
%\authorrunning{S Liu et al.}
%\and Springer Heidelberg, Tiergartenstr. 17, 69121 Heidelberg, Germany
\institute{School of Computer Science and Technology, East China Normal University, China \and Shanghai Transsion Co., Ltd, China \\
\email{shenliu@stu.ecnu.edu.cn},\email{\{mlan,ybwu\}}@cs.ecnu.edu.cn
\email{\{meirong.ma,hao.yuan,jianchao.zhu\}}@transsion.com
}

\maketitle              % typeset the header of the contribution

\begin{abstract}
Pun location is to identify the punning word (usually a word or a phrase that makes the text ambiguous) in a given short text, and pun interpretation is to find out two different meanings of the punning word. Most previous studies adopt limited word senses obtained by WSD(Word Sense Disambiguation) technique or pronunciation information in isolation to address pun location.  For the task of pun interpretation, related work pays attention to various WSD algorithms. In this paper, a model called DANN (\textbf{D}ual-\textbf{A}ttentive \textbf{N}eural \textbf{N}etwork) is proposed for pun location, effectively integrates word senses and pronunciation with context information to address two kinds of pun at the same time. Furthermore, we treat pun interpretation as a classification task and construct pun-gloss pairs as processing data to solve this task. Experiments on the two benchmark datasets show that our proposed methods achieve new state-of-the-art results. Our source code is available in the public code repository\footnote[1]{https://github.com/LawsonAbs/pun}.

\end{abstract}

\section{Introduction}
\begin{figure}[htb]
 % 插入模型结构的文件
  \centering
  \includegraphics[scale=0.32]{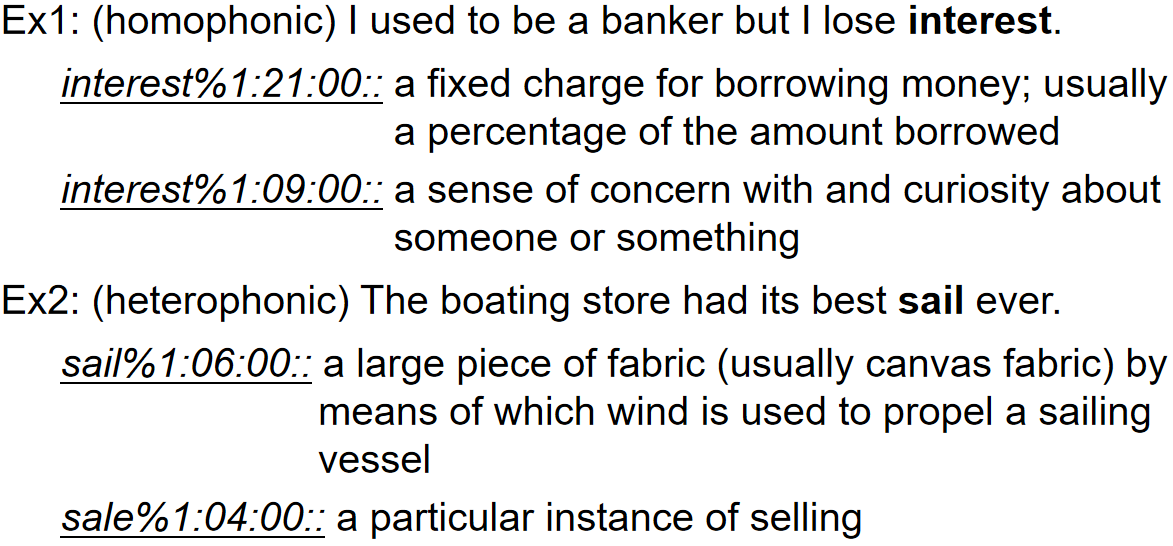}\\  
  \caption {Two samples drawn from two different types puns and their corresponding punning words with the glosses (the definition of word senses) from WordNet \protect\footnotemark[2].}
  \label{fig:two examples}
\end{figure}

\footnote[2]{* means corresponding author}
Puns where the two meanings share the same pronunciation are known as homophonic (i.e., homographic puns), while those relying on similar but not identical-sounding signs are known as heterophonic (i.e., heterographic puns). Figure ~\ref{fig:two examples} shows two examples. Pun location aims to find the word appearing in the text that implies more than one meaning and pun interpretation is an attempt to give the two word senses of the punning word.
% \textit{Ex1: (homographic) I used to be a banker but I lose \textbf{interest}.}
% \textit{Ex2: (heterographic) The boating store has its best \textbf{sail(sale)} ever.}

%\noindent  % 没有缩进
Pun location and interpretation have a wide range of applications \cite{miller-gurevych-2015-automatic,miller-etal-2017-semeval}. Sequence labeling is a general framework to solve pun location \cite{Zou2019JointDA,cai-etal-2018-sense,zhou-etal-2020-boating}. Cai \textit{et al.} \cite{cai-etal-2018-sense} proposed Sense-Aware Neural Model(SAM) which is built on the WSD (Word Sense Ambiguation) algorithms.
%but they only use the limited word senses to conduct pun location. 
It suffers from the bias because of the following reasons: (1) It is inadequate to identify the punning word by using two distinct word senses; (2) The results produced by the WSD algorithms are not always correct, so the error propagation can not be ignored. Moreover, they fail to address the heterographic puns task. Therefore, Zou \textit{et al.} \cite{zhou-etal-2020-boating} add a pronunciation module to the model which is named PCPR(\textbf{P}ronunciation-attentive \textbf{C}ontextualized \textbf{P}un \textbf{R}econgnition) to solve the heterographic puns. However, only utilizing the contextual and pronunciation information, PCPR omits the word senses which are the most important elements in natural language. 
\footnotetext[3]{ https://wordnet.princeton.edu/}
According to the categories of puns, it is intuitive to assume that both word senses and pronunciation are the key points in pun location. So to resolve this problem, we propose a model called DANN(\textbf{D}ual-\textbf{A}ttentive \textbf{N}eural \textbf{N}etwork) to capture the rich semantic and pronunciation information simultaneously. In DANN, the sense-aware and pronunciation-aware modules employ the word meanings and phoneme information respectively. Firstly, unlike SAM, we capture semantic information by paying attention to all meanings of the word automatically rather than selecting several word senses by WSD algorithms in advance.
%Combining word senses definition in WordNet, we put forward a new module to select several optimal word senses.
Secondly, we consolidate word senses, context, and pronunciation information to deal with all kinds of puns.

For pun interpretation, Duluth \cite{Pedersen17} and BuzzSaw \cite{BuzzSaw_2017} both use the WSD algorithm to choose the most probable meaning for the punning word. Specifically, Duluth uses 19 different configurations to create a set of candidate target senses and choose the two most frequent senses from them as the final predicted value. However, one limitation of this approach is the uncertain level of accuracy of the WSD algorithms, which vary from word to word and domain to domain \cite{Pedersen17}. 
%So their model did not work well as the measurements shown in Table ~\ref{tab:experiments_interpretation}. 
Different from Duluth and BuzzSaw, we treat pun interpretation as a sentence pair matching task, that is, we use a pre-training model(e.g., BERT) to select the best matching pun and paraphrase pairs. Concatenating the pun and the gloss of the punning word to one whole sentence, we classify it as yes or no to identify the word sense is correct or not.
%===================== 还需要再添加

In summary, our contributions are as follows:
\noindent
(1) We take full advantage of semantic and phonetic features to conduct the pun location. By the dual-attentive module, both of them can be taken into account.

\noindent
(2) We further explore which meanings of words can lead to rhetorical effects, which is essential for understanding puns. Compared with the simple WSD algorithms, an innovative method through pun-gloss pairs to solve the pun interpretation greatly improves the experimental result.

\noindent
(3) Both models achieve state-of-the-art performance in the benchmark dataset.

\section{Related Work}%============================================================================
\subsection{Pun Location }
%Yang 想干什么？这里简单的罗列是什么意思？
Fixed patterns or characteristics are proposed to solve pun location \cite{miller-etal-2017-semeval,Pedersen17,indurthi-oota-2017-fermi}.  %Then based on the designed sets of features associated with each structure, they constructed different computational classifiers to recognize humor and use Maximal Decrement to extract a minimal set of humor anchors.
%Some work \cite{miller-etal-2017-semeval,Pedersen17,indurthi-oota-2017-fermi} found the following characteristics of puns:(1)A punning word often occurs towards the end of a sentence and has a sense that is semantically related to an earlier word, and the second implication that fits the immediate context in which it occurs;(2)Almost all punning words are content words (i.e., noun, adjective, adverb, verb).
%接着写有哪些论文是从 fix pattern 出发
Yang \textit{et al.} \cite{yang-etal-2015-humor} creatively designed a set of features from four aspects as follows: (a)Incongruity; (b)Ambiguity; (c)Interpersonal Effect; (d) Phonetic Style.
Based on the characteristics of manual design, Duluth \cite{Pedersen17} proposed approaches that relied on WSD and measures of semantic correlation. Using some feature components, Vechtomova \textit{et al.} \cite{vechtomova-2017-uwaterloo} ranked words in the pun by a score calculated as the sum of values of eleven features.
Indurthi \textit{et al.} \cite{indurthi-oota-2017-fermi} select the latter word as a punning word from the maximum similarity word pair. A computational model of humor in puns based on entropy was proposed in \cite{DBLP:conf/cogsci/KaoLG13}. PMI(Pointwise Mutual Information)\cite{DBLP:conf/acl/ChurchH89} to measure the association between words is used in \cite{sevgili-etal-2017-n}.
Doogan \textit{et al.} \cite{DBLP:conf/semeval/DooganGCV17} proposed a probabilistic model to produce candidate words.
Feng \textit{et al.} \cite{DBLP:conf/swisstext/FengSRRB20} first collect 10 kinds of features for this task, then they use logistic regression to find out which word is punning and use the weight of different features to explain why a punning word is detected. 

Based on neural network, some methods are proposed to solve pun location \cite{zhou-etal-2020-boating,cai-etal-2018-sense,Mao2020CompositionalSN}. Mao \textit{et al.} \cite{Mao2020CompositionalSN} proposed CSN-ML (Compositional Semantics Network with Task Learning) to capture the rich semantic information of punning words in a sentence.
%A novel tagging scheme was proposed by \cite{Zou2019JointDA} to address pun detection and pun location jointly.
Cai \textit{et al.} \cite{cai-etal-2018-sense} proposed SAM (Sense-Aware Neural Model) which is built on limited WSD results. Their main idea is modeling multiple sequences of word senses corresponding to different WSD results, which were obtained by various WSD algorithms. 
Zhou \textit{et al.} \cite{zhou-etal-2020-boating} proposed a model named PCPR (Pronunciation-attentive Contextualized Pun Recognition) with current best effectiveness.

Different from these work, we incorporate both semantic and phonetic information into the model and solve pun location perfectly.

\subsection{Pun Interpretation}
%Understanding the meaning of a word is a true challenge. 
Duluth \cite{Pedersen17} use a WSD algorithm on different configurations and then take the MFS(Most Frequent Senses) strategy to predict the appropriate meaning for punning word and get the current best performance. 
%It is a challenge to understand the context. 
However, the MFS strategy is too fixed to address the problem of selecting word senses. Instead of using the WSD algorithm directly, we get the meanings from top-2 pun-gloss pairs with the highest probability as the final results for each target word.

BuzzSaw \cite{BuzzSaw_2017} hypothesize that a pun can be divided into two parts, each containing information about the two distinct senses of the pun, can be exploited for pun interpretation, then they use the method that loosely based on the Lesk algorithm to get the meaning for each polysemous word. Due to error propagation, the pipelined way do not get the best performance in this problem. Therefore, we use pun-gloss pairs to fuse the pun and the gloss of the punning word to one sentence and reduce the process directly. The corresponding experiment shows that our model outperforms all other models.
%Therefore we would get a better outcome.

\section{Methodology}
\begin{figure}[htb]
 % 插入模型结构的文件
  \centering  
  \includegraphics[scale=0.4]{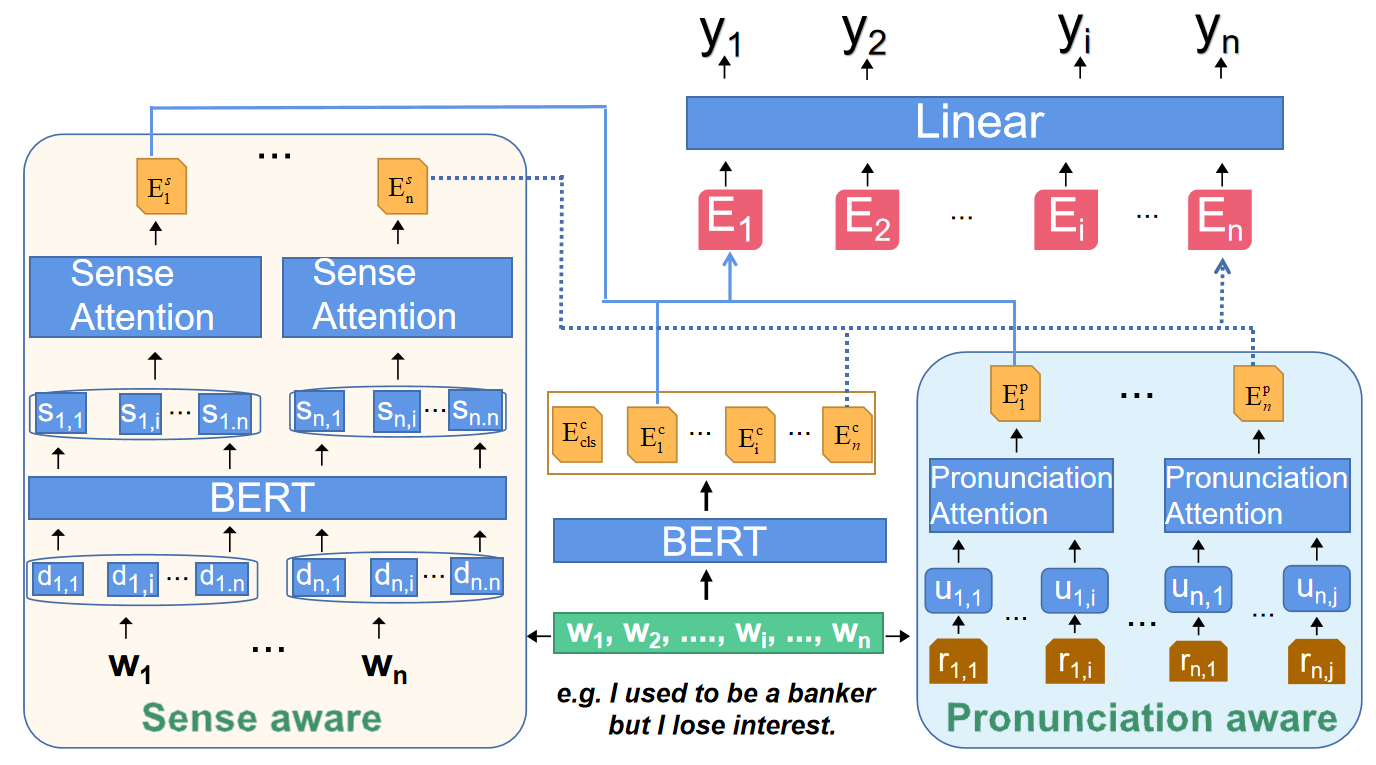}\\  
  \caption{The model architecture of Dual-Attentive Neural Network for Pun Location. We use a dual-attentive module to focus on crucial word senses and pronunciation.}
  \label{fig:architecture_1}
\end{figure}

Figure ~\ref{fig:architecture_1} shows our model architecture for pun location.
Our model is a sequence labeling system, which is based on the adaptation of the BIO annotation ['O', 'P'], where P stands for the punning word tokens and O stands for other tokens. With this tagging scheme, each word in a sentence will be assigned a label. %Our model architecture is showed in Figure ~\ref{fig:architecture}.

Table ~\ref{tab:architecture_2} shows the main construction of training data to solve pun interpretation.
\begin{table}[htb]
%\centering
%\scriptsize
%用于统一调整大小，前者为水平方向，后者为垂直方向
\begin{tabular}{p{120mm}} % 设置表格的宽度
\toprule %添加表格头部粗线
    \textbf{homographic pun:} \\
    I used to be a banker but I lose \underline{interest}. \\
\end{tabular}
\\
\\

\begin{tabular}{llll} %需要13列,所以这里是14
\toprule %添加表格头部粗线
\textbf{Pun-Gloss Pairs of the punning word}& {Label}& {Sense Key}\\
 %有n个&，就表示该行有n+1列
\hline %绘制一条水平横线

{[CLS] I used to be a ...[SEP] a sense of concern ... [SEP]}& Yes & interest\%1:09:00:: \\
{[CLS] I used to be a ...[SEP] a reason for wanting  ... [SEP]}& No & interest\%1:07:01:: \\
{[CLS] I used to be a ...[SEP] excite the curiosity of ... [SEP]}& No & interest\%2:37:00:: \\
{[CLS] I used to be a ...[SEP] a fixed charge for ...[SEP]}&Yes& interest\%1:21:00:: \\
%\cite{xiu-etal-2017-ecnu}
%\bottomrule %添加表格底部粗线
\toprule
\\ %这是必须使用的吗？
\end{tabular}
\caption{
The sample was taken from SemEval-2017 task 7 dataset to explain the construction methods that concatenating the pun and the gloss. The ellipsis "..." indicates the remainder of the sentence.
}
\label{tab:architecture_2}
\end{table}
Inspired by the GlossBERT \cite{GlossBert_2019}, we use the pun-gloss pairs to capture the correlation between the pun and the gloss of target word.
Conventional WSD methods usually return the sense with the highest score. Similarly, we can choose the best and second-best word meanings according to the maximum and sub-maximum probability values returned in the classification process.

%\subsection{Preprocess Work}
%It is well-known that many words have multiple senses. 
\subsection{Pun Location}
\subsubsection{Sense-Aware Module}
The highlight of our model is using the sense-attention module to focus on pertinent word senses automatically.%, which is called self-selecting. 
%In our model, we get word senses through WordNet to help with pun location.

%As we all know, the pun's rhetorical effect is caused by multiple meanings. Even if in heterographic puns, it is the semantics and pronunciation that make a sentence a pun.
%For example, "\textbf{The boating store has its best sail ever.}". This pun is not only because of the pronunciation of the word "sail" and "sale", but also the sense of them. So it is far from enough to only taking pronunciation into account. %The senses of content wordplay a significant role in this task.
%If we could find the specific multiple adaptive senses of the punning word, it would help to judge the word is a punning word or not.
As shown in the lower left corner of the Figure ~\ref{fig:architecture_1}.
Firstly, we get all definitions of the word senses from WordNet for each content word in a pun and denote them as $\{{d_{1,1},...,d_{1,i},...,d_{n,n}\}}$.
%For example, word \emph{interest} have ten kinds of senses in WordNet. Here is one of them, \textit{interest\%1:09:00::} stand for the sense key of the word \textit{interest}:
% \begin{itemize}
%     \item \textit{interest\%1:09:00::} a sense of concern with and curiosity about someone or something
% \end{itemize}
Secondly, we use BERT \cite{DBLP:bert} to process each definition and use its [CLS] token embedding as the representation and denoted them as $\{{s_{1,1},...,s_{1,i},...,s_{1,n}\}}$.
For each word sense embedding $s_{i,j}$ of the word $w_i$, we project $s_{i,j}$  to a trainable vector $s_{i,j}^{'}$ to represent its meaning properties. Based on the word sense embeddings, we apply the attention mechanism \cite{vaswani2017attention} to simultaneously identify important meanings and derive the compositive word sense embedding $E_i^S$.

Specifically, the embedding of word senses are transformed by a fully-connected hidden layer (i.e., $F_S(\cdot)$), and then multiplying by the query vector(i.e., $q$) to measure the importance scores $\alpha_{i,j}$ of word sense embeddings as follows:
\begin{equation}
        v_{i,j} = F_S(s_{i,j})
\end{equation}
\begin{equation}
    \alpha_{i,j} = \frac{v_{i,j}\cdot q} {\sum_k v_{i,k}\cdot q}
\end{equation}
%where $F_S(\cdot)$ is a fully-connected layer with $d_A$ outputs and $d_A$ is the attention size; 
Finally, the synthetical sense embedding $E_i^S$ can be generated by the weighted combination of various embeddings as follows:
\begin{equation}
E_i^S = \sum_j \alpha_{i,j} \cdot s_{i,j}^{'}
\end{equation}

We select context-sensitive paraphrases to help determine whether a word is the target word through using the attention mechanism. Nevertheless, not all words in the input sentence $W_1,W_1,...,W_n$ have the same number of meanings, so this is a hyperparameter, which will be described in \ref{para:hyperparameters}. After that,
a synthetic representation vector(i.e., $E_i^s$) of the various meanings of each word will be got.

\subsubsection{Pronunciation-Aware Module}
%For the sake of fair comparison with the PCPR.
%For heterographic puns, we use the pronunciation module as the same as PCPR. 
It is well-known that pronunciation plays an essential role in the language. Inspired by the PCPR, we also introduce a pronunciation-aware module into DANN to solve the heterographic puns. By projecting pronunciation to the embedding space, words that sound alike are nearby to each other\cite{DBLP:conf/interspeech/BengioH14}. Each word is divided into phonemes(i.e., ${\{r_{1,1},...,r_{1,i},...,r_{n,j}\}}$) which represent the characteristics in pronunciation. Each phoneme is projected to a phoneme embedding space(i.e., $\{{u_{1,1},...,u_{1,i},...,u_{n,j}\}}$). Pronunciation 
vector (i.e.,$E_i^P$) can be obtained with the attention mechanism. Through the pronunciation component, we can join words with the same sound together. %The output of this module for $ith$ word is $E^p_i$.

\subsubsection{Implementation Details}
In our work, we use BERT to get all word embeddings for the whole input sentence. So we can get $E^c,E^s,E^p$ to present contextual embedding, word sense, and pronunciation embedding of the word respectively. Then our model concatenates these embeddings and converts them to a project layer, we can get every word's predicted value $y_i$.

Specifically, first, the BERT model processes the input then gets every word's contextual embedding, we denote them as $E^c$. Second, we use every word's pronunciation embedding, and after the attention process, we get embedding $E^p$ to denote the important pronunciation. Third, word sense embedding serves as the input of the sense-attention module to get the compounded representation of the word, we denote it as embedding $E^s$. 
%$E^{c}_{i}$ is the contextual embedding of the \textit{i-th} word, the pronunciation embedding will be denoted as $E^p_i$, the sense embedding will be denoted as $E^s_i$. 
Last, all embedding parts are concatenated to get the final expression(i.e., $E_i$) of \textit{i-th} word.
\begin{center}
$E_i = E^s_i \oplus E^s_i \oplus E^p_i$    
\end{center}
$E_i$ will be transferred to a project layer to determine whether the \textit{i-th} word is a punning word.

\subsection{Pun Interpretation}
\subsubsection{Framework Overview}
%先介绍为什么用Bert做这个问题
%再概括介绍我们整个任务的framework
BERT uses a "next sentence prediction" task to train text-pair representations, so it can explicitly model the relationship of a pair of texts, which has shown to be beneficial to many pair-wise natural language understanding tasks \cite{DBLP:bert}. To fully leverage gloss information, we construct pun-gloss pairs over puns and all possible senses of the punning word in WordNet, thus treating the WSD task as a sentence-pair classification problem. 

%紧接着介绍整个模型的细节 => 分点介绍
Table ~\ref{tab:architecture_2} shows the main construction process of training examples. 
The sentence containing the punning word is denoted as a \textit{pun} sentence. 
For punning words, we extract glosses of all senses from WordNet.
%详细介绍
An example in homographic pun gives a detailed introduction of the construction method (See Table ~\ref{tab:architecture_2}). \textit{Interest} is a punning word. [SEP] mark is added to the pun-gloss pairs to separate pun from paraphrasing. Each target word has a set of pun-gloss pair training instances with label $\in \{yes, no\}$.

The pun-gloss pairs will serve as inputs to the BERT, and the output of the model is \textit{yes} or \textit{no}. The "\textit{yes}" represents the gloss following the pun is the sense definition of the punning word, the "\textit{no}" stands for the contrary meaning. For clarity and convenience, we use the sense key from WordNet to stand for concrete definition.

\subsubsection{Implementation Details}
We use BERT as our pre-training approach. In training, we get the whole sentence and use BERT to get the [CLS] token embedding, then a linear layer is used to obtain the classification results. 
Cross-entropy loss is used when adjusting model weights during training. When testing, we output the sense key of the punning word with the two maximum probabilities for each pun-gloss pair.

%\section{Experiments}
%Experiments prove that our model achieves new state-of-the-art results. 
\section{Experiment Settings}
\subsection{Dataset and Evaluation Metrics}
We evaluate our models on the SemEval-2017 shared task 7 dataset\footnote[4]{https://alt.qcri.org/semeval2017/task7/}. Homographic puns and heterographic puns have $1607$ and $1271$ samples respectively. Due to the limited data and keep the equity of evaluation, we perform ten cross-validation as the same as PCPR and SAM, then use the average of the evaluation result as the final score. Meanwhile, we use the same metrics with them.

\subsection{Baselines}
\subsubsection{Pun Location}
We compare our model with the following baselines. (1)Olga \cite{vechtomova-2017-uwaterloo}. (2)Idiom Savant \cite{DBLP:conf/semeval/DooganGCV17}. (3)Fermi \cite{indurthi-oota-2017-fermi}. (4)ECNU \cite{xiu-etal-2017-ecnu}. (5)BERT \cite{DBLP:bert}. (6)LRegression \cite{DBLP:conf/swisstext/FengSRRB20}. (7)SAM \cite{cai-etal-2018-sense}. (8)JDL \cite{Zou2019JointDA}. (9)PCPR \cite{zhou-etal-2020-boating}. We directly quote the experimental results of these baselines except BERT.

\subsubsection{Pun Interpretation}
The top-3 competition models in SemEval-2017 task-7 would be used as the baselines.

\label{para:hyperparameters}
\subsection{Hyperparameters}
%We use WordNet to obtain the definition of the word sense. 
Different words have different numbers of meanings, so the number of word senses that should be obtained in the model is a hyperparameter which is denoted as $d_s$. In our work, we use 50 different meanings of a word, and if the word does not have 50 meanings, then it will be initialized to zero embeddings.
%给出系统性能随着definition number变化的趋势
%But there is a balance point between the number of definitions and the performance. So it is important to explore how much definition is optimal. 

\section{Experimental Results and Analysis}
\subsection{Pun Location}
Table ~\ref{tab:experiments metric} shows the specific experimental results. 
\begin{table}[!htb]
%\resizebox{\linewidth}{70pt}{
\centering
 %用于统一调整大小，前者为水平方向，后者为垂直方向
\begin{tabular}{ccccccccccccc} %需要13列,所以这里是14
%\toprule %添加表格头部粗线
\multicolumn{6}{c}{\multirow{2}{*}{System} }& \multicolumn{3}{c}{Homographic}& &\multicolumn{3}{c}{Hetergraphic}\\
\multicolumn{6}{c}{}&\textit{P}&\textit{R}&\textit{F1}&&\textit{P}&\textit{R}&\textit{F1}\\  %有n个&，就表示该行有n+1列
\hline %绘制一条水平横线
%\cite{vechtomova-2017-uwaterloo}
\multicolumn{6}{c}{Olga}& 0.652& 0.652&0.652& &0.797&0.795&0.796\\
%\cite{DBLP:conf/semeval/DooganGCV17}
\multicolumn{6}{c}{Idiom Savant}&0.663& 0.663&0.663& &0.684&0.684&0.684\\
%\cite{indurthi-oota-2017-fermi}
\multicolumn{6}{c}{Fermi}&0.521& 0.521&0.521& &-&-&-\\
%\cite{xiu-etal-2017-ecnu}
\multicolumn{6}{c}{ECNU}&0.337&0.337&0.337& &0.568&0.568&0.568\\
\hline
%\cite{DBLP:BERT}
\multicolumn{6}{c}{BERT*}& 0.884& 0.870&0.877& &0.924&0.925&0.924\\
%\cite{DBLP:conf/swisstext/FengSRRB20}
\multicolumn{6}{c}{LRegression}&0.762&0.762&0.762& &0.849&0.849&0.849\\
%\cite{cai-etal-2018-sense}
\multicolumn{6}{c}{SAM}&0.815&0.747&0.780& &-&-&-\\
%\cite{Zou2019JointDA}
\multicolumn{6}{c}{JDL}&0.835&0.771&0.802& &0.814&0.775&0.794\\
%\cite{zhou-etal-2020-boating}
\multicolumn{6}{c}{PCPR}&\textbf{0.904}&0.875&0.889& &\textbf{0.942}&0.904&0.922\\
\hline
\multicolumn{6}{c}{DANN}&0.895&\textbf{0.914}&\textbf{0.904}& &0.918&\textbf{0.939}&\textbf{0.928}\\
%\bottomrule %添加表格底部粗线
% \\ 这是必须使用的吗？
\end{tabular}
%} % end of resizebox
\caption{
%Performance of our model compared with the top-3 competition results and the methods based  on neural network. * means that the experiments are reproduced in our work, we reference the experimental results from the paper directly otherwise.
Results of DANN and strong baselines on Semeval-2017 task 7 for pun location. * means that the experiments are reproduced in our work.
}
\label{tab:experiments metric}
\end{table}
Compared to PCPR, DANN achieves the highest performance with 1.5\% and 0.6\% improvements of \textit{F1} for the homographic and heterographic datasets respectively. By applying the sense-attention module, we pick out the most valuable meanings to conduct the detecting punning word task. Our model outperforms all baseline models, which indicates that the sense-aware module plays a crucial role, especially in homographic puns. 
%Finally, word senses attention weight explains which word sense contributes to the rhetorical effect.

\subsection{Pun Interpretation}
Table ~\ref{tab:experiments_interpretation} shows that our model achieves the highest performance with 9.16\% improvements of \textit{F1} against the best among the baselines (i.e. Duluth) for the homographic puns.
We posit the reason is that our model makes a good connection between the pun and the gloss of the punning word. So it is possible to see if a relevant definition matches the pun.
\begin{table}[!htb]
%\resizebox{\linewidth}{70pt}{
\centering
 %用于统一调整大小，前者为水平方向，后者为垂直方向
\begin{tabular}{ccccccccc} %需要13列,所以这里是14
%\toprule %添加表格头部粗线
\multicolumn{6}{c}{\multirow{2}{*}{System} }& \multicolumn{3}{c}{Homographic}\\
\multicolumn{6}{c}{}&\textit{P}&\textit{R}&\textit{F1}\\  %有n个&，就表示该行有n+1列
\hline %绘制一条水平横线
\multicolumn{6}{c}{Duluth}& 0.144& 0.168&0.155\\
\multicolumn{6}{c}{BuzzSaw}&0.152& 0.156&0.154\\
\hline
\multicolumn{6}{c}{Ours}&\textbf{0.247}&\textbf{0.247}&\textbf{0.247}\\
%\bottomrule %添加表格底部粗线
% \\ 这是必须使用的吗？
\end{tabular}
%} % end of resizebox
\caption{
%Performance of our model compared with the top-3 competition results and the methods based  on neural network. * means that the experiments are reproduced in our work, we reference the experimental results from the paper directly otherwise.
Results of our model and baselines on Semeval-2017 task 7 for pun interpretation.
}
\label{tab:experiments_interpretation}
\end{table}

\begin{figure}[htb]
\centering
\includegraphics[scale=0.6]{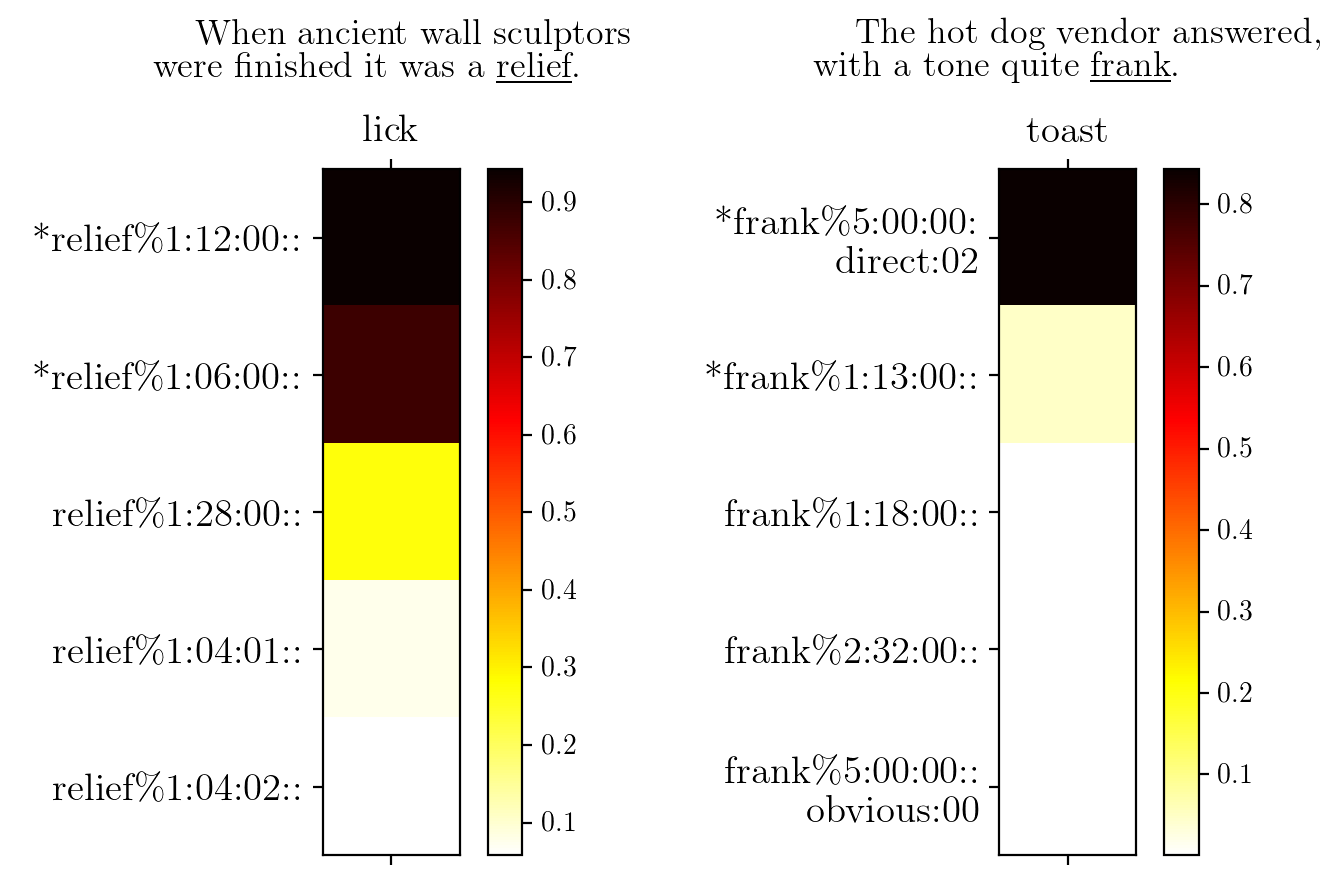}
\caption{The sense key with top-\textit{5} probability for each target word in two samples. The punning words are underlined, and an asterisk indicates the meaning of the word that causes a pun.}
\label{fig:interpretation}
\end{figure}

% It is natural to use the word sense attention scores to explain the word senses that are focused on the pun location process. 
Figure~\ref{fig:interpretation} shows two examples of the explanation in homographic puns.
In the first example, all displaying senses are nouns, \textit{relief\%1:12:00::} and \textit{relief\%1:06:00::} have a higher score because they are closely related to the context. In the second example, although \textit{frank\%5:00:00:direct:02} (adjective) and \textit{frank\%1:13:00::} (noun) have different parts of speech, 
%because of the similar part of speech with the \textit{lick} in the sample text, 
they could also get relatively higher attention scores in this process. We assume that the possible reasons are as follows:
(1) It is easy to find out the primary meaning of \textit{frank}, so the probability of \textit{frank\%5:00:00:direct:02} is the greatest. 
(2) The synonyms of \textit{frank\%1:13:00::} include \textit{hot\_dog\%1:13:01::}. The gloss (i.e., a smooth-textured sausage of minced beef or pork usually smoked; often served on a bread roll) of \textit{frank\%1:13:00::} have a correlation with \textit{hot dog}, so it has the second highest probability score.
%The vast difference between verb and noun is apparent and can explain that our model can utilize the part of speech.

%For heterographic puns, it is first to look for the word that similar in pronunciation with the punning word,  the pronunciation-aware module will find the words with the most similar pronunciation, and the similar word will be in its most primary sense.

\subsection{Analysis}
\subsubsection{Case study}
Table ~\ref{tab:case study} shows the experimental results on several cases between PCPR and DANN. 
% 下面这张表是用来做 case study
\begin{table}[htb]
%\scriptsize
\centering
\begin{tabular}{lllll}
Sentence && PCPR && DANN\\ 
\hline
%\linespread{1.5} 
He stole an invention and then told \textbf{patent} lies. & & lies & & patent\\ 
A thief who stole a calendar \textbf{got} twelve months.& & - && months\\
Finding area is an \textbf{integral} part of calculus. & &  calculus & &  integral\\
%An electric company is looking for high \textbf{energy} employees.& high energy & energy\\
%\multirow{2}{}{}
\hline
\end{tabular}
\caption{\label{citation-guide}
The cases of homographic puns (shown in bold) identified by PCPR and DANN models. 
}\label{tab:case study}
\end{table}
It is obvious to find a significant difference in homographic puns. The valid reason is that the rich semantic information is captured by DANN but forgotten by PCPR. In the first case, \textit{patent} is predicted by the former but \textit{lies} by the latter. We can infer that only considering pronunciation will introduce bias to the model, but the DANN could correct this bias caused by insufficient information through introducing word senses. Except for words with more meanings like \textit{get}, our model got the correct answer on almost every sample. Because these words have so many meanings, it is not a simple matter to find out exactly one definition of them.

\subsubsection{Effect of Number of Word Senses}
\begin{figure}[htb]
 % 插入模型结构的文件
  \centering  
  \includegraphics[scale=0.7]{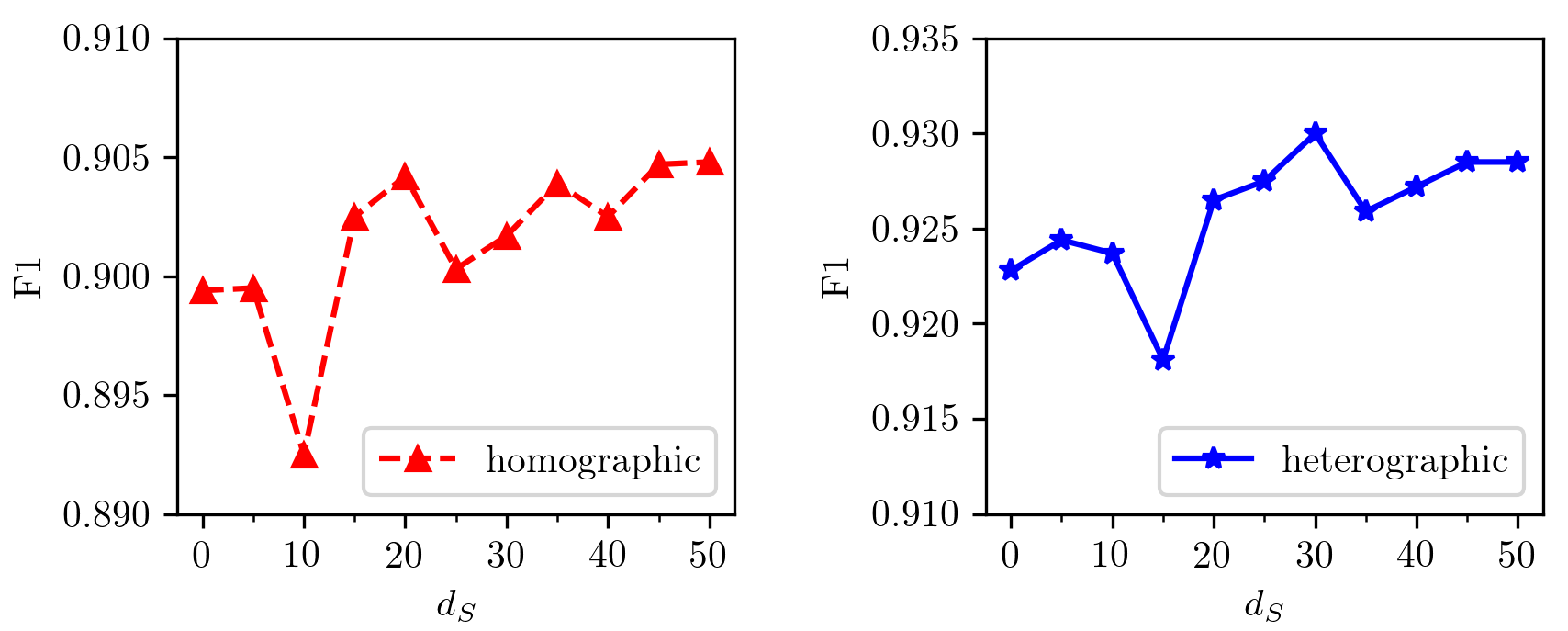}\\
  \caption{Performance over different word sense number in homographic and heterograhic puns.}
  \label{fig:defi_num}
\end{figure}

Figure ~\ref{fig:defi_num} shows the diverse results of the model with a different number of meanings.
There is no doubt that the more word senses you use, the higher the \textit{F1} score you will get. Meanwhile, to keep fair comparison, the hyperparameters we use are exactly the same as in the PCPR, such as phoneme embedding size $d_p$ and attention size $d_A$.

\section{Conclusions and Future Work}
%\section{Conclusions}
In this paper, we propose a novel SOTA model named DANN, which leverages word senses and pronunciation to solve pun location. Empirically, it outperforms previous methods that rely heavily on handcrafted features or another single characteristic. Moreover, we formulate pun interpretation as a classification task and construct pun-gloss pairs to solve it. The experiments show that this method achieves the new best performance with nearly 9.2\% improvement in homographic puns. In the future, we plan to focus on exploring more effective ways to pun interpretation. Furthermore, 
%because puns contain a wealth of emotional information, so they are not only helpful for sentiment analysis, but also for generating humorous text.
due to the rich emotional information in puns, we want to incorporate it into sentiment analysis and text generation to make the machine look smarter.
% the environments 'definition', 'lemma', 'proposition', 'corollary',
% 'remark', and 'example' are defined in the LLNCS documentclass as well.
%

\subsubsection{Acknowledgements.}
We thank the anonymous reviewers for their thoughtful comments. This work has been supported by Shanghai Transsion Co., Ltd.

\bibliographystyle{splncs04}
\bibliography{mybibliography}
%\begin{thebibliography}{10}
%\end{thebibliography}
\end{document}